\documentclass[sigconf,dvipsnames]{acmart}
\usepackage[T1]{fontenc}
\usepackage{graphicx}
\usepackage{hyperref}
\usepackage{amsmath}
\usepackage{booktabs}
\usepackage{longtable}
\usepackage{adjustbox}
\usepackage{array}
\usepackage{soul}
\usepackage{xcolor}
\begin{document}

\title[Leveraging ICL and RAG for Automatic Question Generation in Educational Domains]{Leveraging In-Context Learning and Retrieval-Augmented Generation for Automatic Question Generation in Educational Domains} 

\author{Subhankar Maity}
\orcid{0009-0001-1358-9534}
\affiliation{%
 \institution{IIT Kharagpur}
      \state{West Bengal}
  \country{India}
}
\email{subhankar.ai@kgpian.iitkgp.ac.in}
\authornote{Corresponding Author}

\author{Aniket Deroy}
\orcid{0000-0001-7190-5040}
\affiliation{
 \institution{IIT Kharagpur}
\state{West Bengal}
\country{India}
}
\email{roydanik18@kgpian.iitkgp.ac.in}

\author{Sudeshna Sarkar}
\orcid{0000-0003-3439-4282}
\affiliation{%
 \institution{IIT Kharagpur}
 \state{West Bengal}
 \country{India}
 }
\email{sudeshna@cse.iitkgp.ac.in}

\renewcommand{\shortauthors}{Maity et al.}

\begin{abstract}
Question generation in education is a time-consuming and cognitively demanding task, as it requires creating questions that are both contextually relevant and pedagogically sound. Current automated question generation methods often generate questions that are out of context. In this work, we explore advanced techniques for automated question generation in educational contexts, focusing on In-Context Learning (ICL), Retrieval-Augmented Generation (RAG), and a novel Hybrid Model that merges both methods. We implement GPT-4 for ICL using few-shot examples and BART with a retrieval module for RAG. The Hybrid Model combines RAG and ICL to address these issues and improve question quality. Evaluation is conducted using automated metrics, followed by human evaluation metrics. Our results show that both the ICL approach and the Hybrid Model consistently outperform other methods, including baseline models, by generating more contextually accurate and relevant questions.

\end{abstract}







\maketitle

\section{Introduction}
The demand for personalized and adaptive learning systems in education has grown significantly in recent years \cite{b_0}. Educational technologies, particularly those powered by Artificial Intelligence in Education (AIED), are reshaping how we create learning materials, assess student knowledge, and provide customized instruction \cite{b37}. A crucial element in these systems is Automatic Question Generation (AQG), which refers to the automatic creation of questions from textual content, such as textbook passages, lecture notes, or online educational materials \cite{b38, jan1, jan2, jan3}. AQG has numerous applications, ranging from generating quizzes and assessments to enhancing student engagement through interactive learning platforms \cite{b39}. Despite recent advancements, there remains a significant gap in generating high-quality, contextually relevant, and pedagogically meaningful questions for diverse educational settings \cite{b40}. 

With the advent of neural networks and large language models (LLMs), AQG systems have made significant strides \cite{b45}. Models such as T5 \cite{b8}, BART \cite{b9}, and GPT-3 \cite{b10} have demonstrated remarkable success in tasks such as summarization \cite{b41}, translation \cite{b42}, question-answering \cite{b44}, and dialogue generation \cite{b43}. These models, when fine-tuned on question-answer datasets, can generate more diverse and grammatically correct questions. However, they still face challenges in the educational domain, particularly when it comes to generating pedagogically appropriate questions that align with the instructional goals of a given curriculum \cite{b45}. Additionally, the success of these models often depends on the availability of large, domain-specific training datasets, which are scarce in educational contexts \cite{b45}.

Recent advances in In-Context Learning (ICL) \cite{b10} and Retrieval-Augmented Generation (RAG) \cite{b11} provide promising new avenues for tackling the limitations of traditional AQG methods. Unlike conventional fine-tuning, ICL enables large pre-trained language models to generate task-specific outputs by providing a few input-output examples without updating the model’s parameters. This makes ICL particularly useful in educational settings with limited domain-specific data, where the few-shot examples guide the model to generate contextually relevant questions for new passages. However, the quality of the generated questions relies heavily on the representativeness of the provided examples. On the other hand, RAG, as proposed by \cite{b11}, enhances the generation process by integrating a retrieval mechanism into the generation pipeline. It retrieves semantically relevant external documents and combines them with the input text to generate more informed and contextually rich questions. This is especially beneficial in educational contexts where the provided passage may lack sufficient information, and the retrieval of supplementary resources can add depth and detail to the generated questions. RAG is useful when additional context from sources like textbooks, research papers, or online encyclopedias is needed to create high-quality, domain-specific questions.

Despite their promise, both ICL and RAG have certain limitations when applied independently in the educational AQG domain. ICL is limited by the quality and coverage of the few-shot examples, often generating questions that are too similar to the input examples or fail to address the nuanced details of the input passage. RAG, while effective at incorporating external knowledge, can sometimes retrieve irrelevant or redundant documents \cite{b34}, leading to questions that are either off-topic or overly complicated.

In this paper, we propose that combining the strengths of ICL and RAG can significantly improve the quality of educational question generation. We hypothesize that a hybrid approach, which first retrieves relevant external documents (as in RAG) and then uses a few-shot learning mechanism (as in ICL) to guide question generation, will result in questions that are not only contextually relevant but also pedagogically meaningful. By leveraging external information through retrieval and guiding the model’s output with carefully selected examples, we aim to address the shortcomings of each individual method and generate a broader range of questions that target various cognitive levels in educational assessments.

This paper makes several key contributions to the field of AQG, particularly in the context of educational applications:

\begin{itemize}
    
    \item A comparative analysis of In-Context Learning (ICL) and Retrieval-Augmented Generation (RAG) for automated question generation in the educational domain. We evaluate how each method performs individually in generating questions from textbook passages and assess their respective strengths and weaknesses.

   \item  A novel hybrid model that combines ICL and RAG to generate higher-quality, contextually accurate, and pedagogically aligned questions. Our hybrid approach leverages external knowledge through retrieval and incorporates in-context learning to guide the generation process.

   \item A comprehensive evaluation of the proposed methods using both automated metrics (e.g., BLEU-4, ROUGE-L, METEOR, ChRF, BERTScore) and human evaluation by educators. We assess the \textit{grammaticality}, \textit{appropriateness}, \textit{relevance}, \textit{complexity}, and \textit{answerability}, providing insights into the pedagogical value of each approach.

\end{itemize}

The rest of the paper is structured as follows: Section \ref{sec2} reviews related work in AQG, ICL, and RAG. Section \ref{sec3} outlines the task definition and mathematical formulation of the problem. Section \ref{sec4} discusses the dataset used in our study. In Section \ref{sec5}, we describe the methodology, detailing the implementation of ICL, RAG, and our proposed hybrid model. The experimental setup is presented in Section \ref{sec6}. Section \ref{sec7} covers the automated and human evaluation metrics used in this study. Results and analysis are presented in Section \ref{sec8}, where we compare the models using both automatic and human evaluations and provide a detailed analysis of the results. Finally, Section \ref{sec9} concludes the paper and discusses potential future work.


\section{Related Work}\label{sec2}

\subsection{Automatic Question Generation (AQG)}

The field of Automatic Question Generation (AQG) has evolved from early rule-based systems to more sophisticated neural network-based approaches. Initial research in AQG primarily focused on template-based methods \cite{b20, b21, b22}, which generate questions by applying syntactic and semantic rules to predefined templates. For example, \cite{b19} used syntactic transformations to generate factual questions from declarative sentences. However, template-based methods such as \cite{b20} lack flexibility and are highly dependent on predefined rules, which limit their applicability to different domains and complex question types.

With the rise of neural networks, sequence-to-sequence (Seq2Seq) models such as T5 \cite{b8} and BART \cite{b9} have been successfully applied to AQG \cite{b7, b12, b23}. These models are trained end-to-end to map a given passage to a corresponding question, generating more diverse and grammatically correct questions. However, neural models often struggle with domain-specific knowledge, especially when training data is limited. Moreover, they tend to generate questions that are either too generic or not pedagogically aligned with the content's educational goals.

\subsection{In-Context Learning (ICL)}

In-Context Learning (ICL) was introduced by \cite{b10} as part of the capabilities of the GPT-3 language model. Unlike fine-tuning approaches, where a model is trained on a specific task using a large dataset, ICL allows large language models to perform a task based on a few examples provided as input \cite{b24}. The model processes these examples and generalizes them to generate appropriate outputs for unseen instances. ICL has demonstrated success in various tasks, including text classification \cite{b25}, translation \cite{b26}, and summarization \cite{b27, b28}, making it particularly valuable in situations where data is scarce.

However, in the context of AQG, ICL’s performance has not been extensively studied. The challenge lies in crafting effective few-shot prompts that guide the model toward generating pedagogically meaningful questions. Existing work has shown that ICL is sensitive to prompt design, and small changes in the examples provided can lead to significantly different outputs \cite{b24}. Furthermore, ICL's reliance on in-context examples means it may fail to capture deeper contextual information that is not present in the input passage but is essential for generating complex educational questions.

\subsection{Retrieval-Augmented Generation (RAG)}

Retrieval-Augmented Generation (RAG), introduced by \cite{b11}, addresses the limitations of purely generative models by incorporating an external retrieval mechanism. In a typical RAG framework, the model first retrieves relevant documents from an external corpus based on the input text \cite{b29}. The retrieved documents provide additional context, which is used to generate more accurate and contextually enriched outputs \cite{b30}. This approach is particularly beneficial for knowledge-intensive tasks where the input text alone does not provide sufficient information.

RAG has been applied successfully to various tasks, such as knowledge-based question answering \cite{b31, b32, b33} and code summarization \cite{b35}. Nevertheless, RAG's performance in the context of AQG has not been widely explored. In the context of AQG, RAG can retrieve supplementary information from external sources, such as textbooks or research papers, to generate questions that are not only contextually aligned with the input passage but also pedagogically relevant. However, the retrieval process may introduce noise if irrelevant or redundant documents are retrieved \cite{b34}, which can negatively impact the quality of the generated questions.

\section{Task Definition}\label{sec3}

\subsection{Problem Statement}

The task of \textit{Automatic Question Generation (AQG)} can be formulated as follows: given an input passage \( P \), the objective is to generate a question \( Q \) that is relevant to the content of \( P \), contextually accurate, and aligned with educational goals. 

\subsection{Input and Output Representation}

The input to the model is a passage \( P \), which can be a sentence, a paragraph, or a longer text excerpt from educational material. The output is a question \( Q \) that is relevant to \( P \) and suitable for use in an educational setting. Formally, we define the task as learning a function:
\[
f(P) = Q
\]

where \( P \) is the input passage, and \( Q \) is the generated question.

\subsection{In-Context Learning (ICL)}

In the \textit{In-Context Learning (ICL)} paradigm, given an input passage \( P_{\text{new}} \) and a set of \( k \) few-shot examples \( \{(P_1, Q_1), (P_2, Q_2), \dots, (P_k, Q_k)\} \), the model generates a new question \( Q_{\text{new}} \) corresponding to \( P_{\text{new}} \):
\[
Q_{\text{new}} = f_{\text{ICL}}(P_{\text{new}}, \{(P_i, Q_i)\}_{i=1}^k)
\]

Here, the few-shot examples serve as prompts to guide the question generation process for the new passage.

\subsection{Retrieval-Augmented Generation (RAG)}

For \textit{Retrieval-Augmented Generation (RAG)}, the task is extended by incorporating an external retrieval mechanism. Given a passage \( P \), the model retrieves a set of relevant documents \( \{R_1, R_2, \dots, R_k\} \) from an external corpus. These documents provide additional context, and the final question \( Q \) is generated as:
\[
Q = f_{\text{RAG}}(P, \{R_i\}_{i=1}^k)
\]

\subsection{Hybrid Model}

Our proposed \textit{Hybrid Model} combines the advantages of both ICL and RAG. The model first retrieves a set of documents \( \{R_i\}_{i=1}^k \) for the input passage \( P \), and then uses few-shot learning to generate the question \( Q \) based on both the passage and retrieved documents:
\[
Q = f_{\text{Hybrid}}(P, \{R_i\}_{i=1}^k, \{(P_i, Q_i)\}_{i=1}^m)
\]

Here, the retrieval step enriches the context for question generation, while the few-shot examples (i.e., $m$ examples) help guide the model towards generating pedagogically relevant questions.

\section{Dataset} \label{sec4}
We used the EduProbe dataset \cite{b7}, which comprises 3,502 question-answer pairs across various subjects: 858 pairs related to History, 861 pairs related to Geography, 802 pairs related to Economics, 606 pairs related to Environmental Studies, and 375 pairs related to Science. The dataset was curated from segments of varying lengths, extracted from a diverse range of chapters in National Council of Educational Research and Training (NCERT)\footnote{\url{https://en.wikipedia.org/wiki/National_Council_of_Educational_Research_and_Training}} textbooks across several subjects, covering standards from $6^{th}$ to $12^{th}$. Each entry in the dataset includes a context (or passage), a long prompt, a short prompt, and a question. For our experiments, we extracted only the context (or passage) and question, focusing on evaluating how well the models generate questions based on the provided contexts (or passages). We used the same training and test datasets as in \cite{b7}.

\section{Methodology}\label{sec5}

In this section, we describe the methodologies employed for question generation using In-Context Learning (ICL), Retrieval-Augmented Generation (RAG), and a Hybrid Model that combines both approaches. 

\subsection{In-Context Learning (ICL) Approach}

For the \textit{ICL} approach, we use the GPT-4 \cite{b46} model to generate questions based on a few-shot prompt. Each prompt consists of \( k \) example input-output pairs \( \{(P_1, Q_1), \dots, (P_k, Q_k)\} \), where each pair consists of a passage and a corresponding question. Given a new passage \( P_{\text{new}} \), the model generates a question \( Q_{\text{new}} \) using the few-shot examples as context. The general structure of the ICL prompt is as follows:
\[
\langle P_1, Q_1 \rangle, \langle P_2, Q_2 \rangle, \dots, P_{\text{new}} \rightarrow Q_{\text{new}}
\]

We experimented with different numbers of examples \( k \) and optimized the structure of the prompt for maximum performance.

\subsection{Retrieval-Augmented Generation (RAG) Approach}

The \textit{RAG} model employs the BART architecture as its generative backbone, further refined by a retrieval module. This retrieval system, implemented using \textit{FAISS} \cite{b1}, searches through an extensive external corpus of educational materials\footnote{\url{https://ncert.nic.in/textbook.php}}
specifically tailored for school-level subjects such as History, Geography, Economics, Science, and Environmental Studies. For a given passage \( P \), the system identifies the most relevant documents from this curated corpus. These retrieved documents denoted as \( \{R_i\}_{i=1}^k \), are concatenated with \( P \), and the combined input is processed through a fine-tuned BART model trained on the EduProbe training set \cite{b7} to generate a question \( Q \). The fine-tuning of BART on EduProbe ensures that the model is adept at generating questions based on educational content, further enhancing its performance. Formally, question generation in RAG is defined as:
\[
Q = f_{\text{RAG}}(P, \{R_i\}_{i=1}^k)
\]

The retrieval module enriches the context, allowing the model to generate questions that are better aligned with the content and objectives of the passage.

\subsection{Hybrid Model}

Our \textit{Hybrid Model} combines the retrieval-based context enrichment of RAG with the few-shot learning mechanism of ICL using GPT-4. First, relevant documents are retrieved for a given passage \( P \). Then, the few-shot learning mechanism uses these retrieved documents alongside the input passage to generate a more contextually accurate and pedagogically meaningful question. The hybrid approach can be mathematically defined as:
\[
Q = f_{\text{Hybrid}}(P, \{R_i\}_{i=1}^k, \{(P_i, Q_i)\}_{i=1}^m)
\]

Here, \( P \) is the input passage, \( \{R_i\}_{i=1}^k \) are the retrieved documents, and \( \{(P_i, Q_i)\}_{i=1}^m \) are the few-shot examples used to guide the question generation process.

\section{Experimental Setup} \label{sec6}
Our experiments evaluate the effectiveness of the In-Context Learning (ICL), Retrieval-Augmented Generation (RAG), and Hybrid Model for Automatic Question Generation (AQG). We aim to assess and compare these models' performance using a variety of automated and human evaluation metrics.

\subsection{Baseline Models}

Following \cite{b12, b7}, we fine-tune the best-performing models (based on automated evaluation), such as the T5-large and BART-large architectures, on the EduProbe training dataset \cite{b7}. The T5-large model, implemented from the Hugging Face Transformers library\footnote{\url{https://huggingface.co/models}}, uses a sequence-to-sequence framework with attention mechanisms to align passages with questions, providing a robust baseline for comparison. Similarly, the BART-large model, also from Hugging Face, employs a transformer encoder-decoder structure to generate questions from passages, serving as another strong baseline for our evaluation.

\subsection{In-Context Learning (ICL)}

We implemented ICL using GPT-4\footnote{\url{https://platform.openai.com/docs/models/gpt-4-turbo-and-gpt-4}}, providing it with different few-shot settings ($k=3,5,7$). In this setup, the model uses a set of example passage-question pairs to generate questions for new passages. We tested various numbers of examples to determine the optimal few-shot settings for generating relevant and coherent questions.

\subsection{Retrieval-Augmented Generation (RAG)} 

For RAG, we utilized BART-large\footnote{\url{https://huggingface.co/facebook/bart-large}} as the backbone model and integrated a retrieval module with FAISS for efficient document retrieval. For each input passage, the retrieval module identifies and fetches the top $ k=5 $ relevant documents from a large corpus of educational material. These documents are concatenated with the input passage, and BART-large generates questions based on the combined context.

\subsection{Hybrid Model}

The Hybrid Model combines retrieval and in-context learning techniques. We first retrieve $k=5$ relevant documents for each passage, then use a few-shot prompt with $m=5$ examples to generate questions. This model leverages both the additional context from retrieved documents and the few-shot learning capabilities to produce high-quality questions.

\section{Evaluation Metrics} \label{sec7}

The performance of the models was evaluated using several automated evaluation metrics: BLEU-4 \cite{b3}, ROUGE-L \cite{b2}, METEOR \cite{b4}, ChRF \cite{b5}, and BERTScore \cite{b6}. We use the implementations of these metrics provided by the SummEval package\footnote{\url{https://github.com/Yale-LILY/SummEval}}.

Acknowledging the limitations of automated metrics in text generation research \cite{b15, b16, b17, b18}, we conducted a human evaluation involving three high school teachers and two high school students. Each evaluator reviewed a total of 1,400 questions across seven settings: T5-large (baseline), BART-large (baseline), ICL ($k=3, 5, 7$), RAG ($k=5$), and Hybrid model ($k=5, m=5$). They rated each question on a scale from 1 (worst) to 5 (best) based on five criteria: \textit{Grammaticality} \cite{b7, b13, b14} (correctness of grammar independent of context), \textit{Appropriateness} \cite{b7} (semantic correctness irrespective of context), \textit{Relevance} \cite{b7} (alignment with the given context), \textit{Complexity} \cite{b7} (level of reasoning required to answer), and \textit{Answerability} \cite{b13, b14} (whether the question can be answered from the provided context).

To evaluate the level of agreement among the five raters for each generated question, we use Fleiss’s kappa as the measure of inter-rater agreement. The resulting kappa scores are 0.51, 0.48, 0.45, 0.45, and 0.49 for \textit{grammaticality}, \textit{appropriateness}, \textit{relevance}, \textit{complexity}, and \textit{answerability}, respectively. These kappa values suggest a moderate level of agreement across all human evaluation metrics \cite{b36}.

\begin{table*}[ht]
\centering
\caption{Automatic evaluation results for different models. The highest value for each metric, achieved by any model, is highlighted in \textcolor{blue}{blue}, and values marked with * are statistically significant based on student's $t$-test at the 95\% confidence interval compared to the lowest corresponding baseline value.}
\label{tab:automatic_metrics}
\renewcommand{\arraystretch}{1}
\begin{tabular}{l|c|c|c|c|c}
\toprule
\toprule
\textbf{Model} & \textbf{BLEU-4} & \textbf{ROUGE-L} & \textbf{METEOR} & \textbf{ChRF} & \textbf{BERTScore} \\
\midrule
\midrule
T5-large (Baseline) \cite{b7} & 21.59 & 53.90 & 32.20 & 57.03 & 71.80 \\
BART-large (Baseline) \cite{b7} & 20.05 & 51.60 & 31.90 & 54.96 & 74.20 \\ \midrule
ICL ($k=3$) & 22.65 & 54.24 & 32.98 & 58.47 & 74.93 \\
ICL ($k=5$) & \textcolor{blue}{22.87} & 54.84 & 33.58 & 59.42* & 75.60* \\
ICL ($k=7$) & 22.69  & \textcolor{blue}{55.95}* & \textcolor{blue}{34.62} & \textcolor{blue}{60.48}* & \textcolor{blue}{75.92}* \\ \midrule
RAG ($k=5$) & 20.76 & 52.60 & 32.07 & 56.93 & 70.20  \\ \midrule
Hybrid Model ($k=5, m=5$) & 21.45 & 53.79 & 33.69 & 57.78 & 71.45 \\
\bottomrule
\bottomrule
\end{tabular}

\end{table*}

\begin{table*}[ht]
\centering
\caption{Human evaluation results for generated questions across different models on \textit{grammaticality} (Gramm), \textit{appropriateness} (Appr), \textit{relevance} (Rel), \textit{complexity} (Comp), and \textit{answerability} (Answ). The highest value for each metric, achieved by any model, is highlighted in \textcolor{blue}{blue}, and values marked with * are statistically significant based on student's $t$-test at the 95\% confidence interval compared to the lowest corresponding baseline value.}
\label{tab:human_evaluation}
\renewcommand{\arraystretch}{1}
\begin{tabular}{l|c|c|c|c|c}
\toprule
\toprule
\textbf{Model} & \textbf{Gramm} & \textbf{Appr} & \textbf{Rel} & \textbf{Comp} & \textbf{Answ} \\
\midrule
\midrule
T5-large (Baseline) \cite{b7} & 4.65 & 4.45 & 3.92 & 3.57 & 3.21 \\
BART-large (Baseline) \cite{b7} & 3.81 & 3.98 & 3.60 & 3.60 & 3.15 \\ \midrule
ICL ($k=3$) & 4.67* & 4.50* & 3.97* & 3.65 & 3.20 \\
ICL ($k=5$) & 4.72* &  4.56* & 4.03* & 3.78 & 3.24 \\
ICL ($k=7$) & 4.76* & 4.62* & 4.08* & 3.84 & \textcolor{blue}{3.31} \\ \midrule
RAG ($k=5$) & 3.90 & 4.10 & 3.70 & 3.74 & 2.90\\ \midrule
Hybrid Model ($k=5, m=5$) & \textcolor{blue}{4.84}* & \textcolor{blue}{4.74}* & \textcolor{blue}{4.25}* & \textcolor{blue}{4.02}* & 3.20 \\
\bottomrule
\bottomrule
\end{tabular}

\end{table*}

\begin{table*}[ht]
\centering
\caption{A sample (context and gold standard question) from the History subject of the EduProbe dataset \cite{b7}, including questions generated by various models and settings. The \hl{highlighted} color indicates \textit{\underline{irrelevant}} or \textit{\underline{unanswerable}} questions according to the context.}
\resizebox{\textwidth}{!}{%
\begin{tabular}{p{0.9\textwidth}}
\toprule
\textbf{Context:} During the medieval period in India, Islamic rulers held significant power, leading to the blending of Indian and Islamic cultures, which can still be observed in the architecture and artwork created at that time. The country was governed and administered by notable rulers who made exceptional contributions in the fields of art, culture, literature, and architecture during this period. \\ 
\textbf{Gold Standard Question:} What is the contribution of the medieval period to Indian history? \\ \midrule
\end{tabular}%
}
\vspace{1em} 
\resizebox{\textwidth}{!}{%
\begin{tabular}{l|p{0.8\textwidth}}
 \midrule
\textbf{Model} & \textbf{Generated Question} \\  \midrule
T5-large (Baseline) \cite{b7} & What happened in the medieval period in India, which saw a strong control of Islamic rulers? \\  \midrule
BART-large (Baseline) \cite{b7} & \hl{What was the main difference between the medieval period in India and the Chalcolithic period?} \\  \midrule
ICL ($k=3$) & What were the major cultural influences of the Islamic rulers during India's medieval period? \\  \midrule
ICL ($k=5$) & How did Islamic rulers contribute to the architectural development in India during the medieval era? \\  \midrule
ICL ($k=7$) & What were the key impacts of the medieval Islamic rulers on Indian art and architecture, and how are these contributions reflected today? \\  \midrule
RAG ($k=5$) & How did medieval Indian rulers influence the country's cultural and artistic heritage? \\  \midrule
Hybrid Model ($k=5, m=5$) & In what ways did the Islamic rulers of medieval India shape the cultural and architectural landscape, and what are some prominent examples of their contributions? \\ \bottomrule
\end{tabular}%
}
\label{ex1}
\end{table*}

\begin{table*}[ht]
\centering
\caption{A sample (context and gold standard question) from the Economics subject of the EduProbe dataset \cite{b7}, including questions generated by various models and settings. The \hl{highlighted} color indicates \textit{\underline{irrelevant}} or \textit{\underline{unanswerable}} questions according to the context.}
\resizebox{\textwidth}{!}{%
\begin{tabular}{p{0.9\textwidth}}
\toprule
\textbf{Context:} Purchasing power parity (PPP) is an economic indicator that signifies the purchasing power of the currencies of various nations of the world against each other. It helps in comparing living standards between different countries and estimating economic productivity. \\ 
\textbf{Gold Standard Question:} What does purchasing power parity do? \\ \midrule
\end{tabular}%
}
\vspace{1em} 
\resizebox{\textwidth}{!}{%
\begin{tabular}{l|p{0.8\textwidth}}
\midrule
\textbf{Model} & \textbf{Generated Question} \\ \midrule
T5-large (Baseline) \cite{b7} & \hl{What is the importance of a country’s development index? How does it help the poor?} \\ \midrule
BART-large (Baseline) \cite{b7} & \hl{What is the importance of ease of living standards in simple words, and why is it important?} \\ \midrule
ICL ($k=3$) & What is the function of purchasing power parity (PPP) in comparing currencies? \\ \midrule
ICL ($k=5$) & How does purchasing power parity (PPP) help in comparing living standards between countries? \\ \midrule
ICL ($k=7$) & What role does purchasing power parity (PPP) play in comparing economic productivity between different countries? \\ \midrule
RAG ($k=5$) & What is the purpose of purchasing power parity (PPP) in assessing economic productivity? \\ \midrule
Hybrid Model ($k=5, m=5$) & How does purchasing power parity (PPP) assist in comparing living standards and currencies across nations? \\ \bottomrule
\end{tabular}%
}
\label{ex2}
\end{table*}

\section{Results and Analysis} \label{sec8}

In the automated evaluation (see Table \ref{tab:automatic_metrics}), In-Context Learning (ICL) with $k=7$ demonstrates the best overall performance, surpassing other configurations such as RAG ($k=5$) and the Hybrid Model ($k=5, m=5$). ICL with $k=7$ excels in ROUGE-L, METEOR, CHrF, and BERTScore, while ICL with $k=5$ achieves the highest BLEU-4 score. This advantage can be attributed to ICL's ability to effectively leverage multiple examples to generate more contextually relevant questions. In contrast, RAG and the Hybrid Model depend on external document retrieval, which sometimes retrieves content misaligned with the input passage, reducing their effectiveness in generating closely related questions. Nevertheless, both RAG and the Hybrid Model consistently outperform the fine-tuned baseline models (i.e., T5-large and BART-large) across all automated evaluation metrics, demonstrating their overall superiority in question generation tasks. The fine-tuned models, while reliable, lack the additional context provided by retrieval or few-shot examples, making them less effective than the more advanced techniques (i.e., ICL, RAG or Hybrid Models).

As shown in Table \ref{tab:human_evaluation}, the human evaluation results indicate that the Hybrid Model ($k = 5, m = 5$) consistently outperforms other models across key metrics such as \textit{grammaticality}, \textit{appropriateness}, \textit{relevance}, and \textit{complexity}. The Hybrid Model achieves the highest scores in most human evaluation metrics, suggesting its ability to generate more linguistically correct, contextually appropriate, and pedagogically complex questions. In terms of \textit{relevance} and \textit{complexity}, the Hybrid Model demonstrates a clear advantage, likely due to its integration of both retrieval and few-shot learning techniques, which allows it to create questions that are well-aligned with the passage and involve deeper reasoning. This is followed by the ICL with $k=7$, which also performs strongly across all metrics, particularly in \textit{grammaticality}, \textit{appropriateness}, and \textit{relevance}, but slightly lags behind the Hybrid Model in \textit{complexity}. However, ICL with $k=7$ outperforms other models in generating more \textit{answerable} questions, as it leverages multiple examples to better align the questions with the context, making them more straightforward to answer. RAG ($k=5$), however, shows weaker performance, particularly in \textit{answerability}, where it scores lower compared to both the ICL and Hybrid models. This could be attributed to its reliance on external document retrieval, which may introduce content that is less directly tied to the passage, thus affecting the \textit{relevance} and \textit{answerability} of the generated questions. The fine-tuned baseline models, T5-large and BART-large, perform adequately but fall behind the more advanced models. T5-large exhibits a stronger performance compared to BART-large across most metrics, particularly in \textit{grammaticality} and \textit{appropriateness}, but neither model matches the performance of the ICL or Hybrid approaches. These findings underscore the benefits of incorporating retrieval and few-shot examples for generating more contextually relevant and complex educational questions. 

Tables \ref{ex1} and \ref{ex2} show two data samples (context and gold standard question) from the EduProbe dataset \cite{b7}, covering History and Economics. They also display the corresponding questions generated by T5-large (baseline), BART-large (baseline), ICL ($k = 3, 5, 7$), RAG ($k = 5$), and the Hybrid model ($k = 5, m = 5$). It can be observed that while baseline models sometimes produce questions that are \textit{irrelevant} or \textit{unanswerable} from the context, the proposed techniques, particularly ICL, generate questions that are more closely aligned with the provided context.

\section{Conclusions and Future Work} \label{sec9}
In this study, we explored advanced methods for automated question generation in educational contexts, focusing on In-Context Learning (ICL), Retrieval-Augmented Generation (RAG), and a novel Hybrid Model that combines both techniques. Our results reveal that ICL with $k=7$ demonstrates superior performance in automated evaluation metrics, such as ROUGE-L, METEOR, CHrF, and BERTScore. On the other hand, the Hybrid Model ($k=5, m=5$) shows superior performance in human evaluation metrics, including \textit{grammaticality}, \textit{appropriateness}, \textit{relevance}, and \textit{complexity}. The combination of retrieval and in-context learning techniques allows the Hybrid Model to generate questions with greater depth and alignment with educational contexts. Overall, both ICL and the Hybrid Model significantly outperform the previous baseline models, T5-large and BART-large, in both automated and human evaluations. These advanced models surpass baselines in generating more relevant and contextually appropriate questions, demonstrating their enhanced capabilities in generating educational questions.

In future work, we plan to expand evaluations to diverse datasets and educational domains to better assess the generalizability of our models. We also aim to incorporate feedback from educators and students to refine our approaches, ensuring they meet curriculum standards and effectively support learning outcomes. In addition, we plan to explore other state-of-the-art LLMs, such as Gemini and Llama-2-70B, for in-context learning. These initiatives will improve the development of more effective and contextually accurate AQG systems.

\section*{Acknowledgments}
The first two authors gratefully acknowledge the Ministry of Human Resource Development (MHRD), Government of India, for funding their doctoral research. The authors also thank the anonymous reviewers for their valuable feedback and insightful suggestions, which significantly enhanced this work. Additionally, they acknowledge the use of \href{https://chatgpt.com/}{ChatGPT} for proofreading, polishing, and enhancing the clarity and style of their initial drafts, as well as for reviewing the final manuscript prior to submission.
\bibliographystyle{ACM-Reference-Format}
\bibliography{sample-base}









\end{document}